\documentclass{article}

\usepackage{microtype}
\usepackage{graphicx}
\usepackage{subcaption}
\usepackage{booktabs} 
\usepackage{multirow}            

\usepackage{hyperref}

\usepackage[preprint]{icml2026}

\usepackage{amsmath}
\usepackage{amssymb}
\usepackage{mathtools}
\usepackage{amsthm}
\usepackage{siunitx}
\usepackage{nicefrac}
\usepackage{orcidlink}

\usepackage[capitalize,noabbrev]{cleveref}

\setcitestyle{numbers,square,comma,sort&compress}

\usepackage{tikz}
\PassOptionsToPackage{RGB}{xcolor}

\definecolor{custom2}{RGB}{222, 143, 5}
\definecolor{custom3}{RGB}{2, 158, 115}
\definecolor{custom4}{RGB}{213, 94, 0}

\theoremstyle{plain}

\theoremstyle{definition}

\theoremstyle{remark}

\usepackage[textsize=tiny]{todonotes}

\icmltitlerunning{Deep Graph Learning will stall without Network Science}

\begin{document}

\twocolumn[
  \icmltitle{Deep Graph Learning will stall without Network Science}




  \begin{icmlauthorlist}
    \icmlauthor{Christopher Bl{\"o}cker~\orcidlink{0000-0001-7881-2496}}{jmu,umucs}
    \icmlauthor{Martin Rosvall~\orcidlink{0000-0002-7181-9940}}{umu}
    \icmlauthor{Ingo Scholtes~\orcidlink{0000-0003-2253-0216}}{jmu}
    \icmlauthor{Jevin D.\ West~\orcidlink{0000-0002-4118-0322}}{uwa}
  \end{icmlauthorlist}

  \icmlaffiliation{jmu}{Chair of Machine Learning for Complex Networks, Center for Artificial Intelligence and Data Science (CAIDAS), Julius-Maximilians-Universi{\"a}t W{\"u}rzburg, Germany}
  \icmlaffiliation{umucs}{Department of Computing Science, Ume{\aa} Universitet, Ume{\aa}, Sweden}
  \icmlaffiliation{umu}{Integrated Science Lab (IceLab), Department of Physics, Ume{\aa} Universitet, Ume{\aa}, Sweden}
  \icmlaffiliation{uwa}{Center for an Informed Public, Information School, University of Washington, Seattle, USA}

  \icmlcorrespondingauthor{Christopher Bl{\"o}cker}{christopher.bloecker@uni-wuerzburg.de}

  \icmlkeywords{Machine Learning, ICML}

  \vskip 0.3in
]



\printAffiliationsAndNotice{}  

\begin{abstract}
Deep graph learning focuses on flexible and generalizable models that learn patterns in an auto\-mated fashion.
Network science focuses on models and measures revealing the organizational principles of complex systems with explicit assumptions.
Both fields share the same goal: to better model and understand patterns in graph-structured data.
However, deep graph learning prioritizes empirical performance but ignores fundamental insights from network science.
\emph{Our position is that deep graph learning will stall without insights from network science}.
In this position paper, we formulate six Calls for Action to leverage untapped insights from network science to address current issues in deep graph learning, ensuring the field continues to make progress.
\end{abstract}

\section{Introduction}\label{sec:intro}
In 1982, John Hopfield introduced a neural network model that sparked a flurry of innovations: content-addressable memory, energy dynamics, error correction, and nonlinear architecture \cite{hopfield1982neural}.
The Nobel Prize committee recently recognized the role these innovations played in the development of modern machine learning\footnote{\url{https://www.nobelprize.org/prizes/physics/2024/hopfield/facts/}}.
Less recognized, the paper influenced network science just as profoundly.
The Hopfield network demonstrated the critical connection between network topology and the collective behaviour of complex systems---one of the enduring themes of network science and now one of the central challenges in deep graph learning.
The two fields have diverged since Hopfield's paper.
We see a need for that to change, and argue for integrating network science insights into deep graph learning.

Deep graph learning faces several core challenges. Methods must augment data to cope with limited training data. They must pool node representations for graph-level learning.
They must incorporate the time dimension of temporal graphs.
And they must develop message-passing schemes that incorporate higher-order interactions beyond pairwise edges.
Network science has been thinking about these issues for years, though from a different perspectives and with different motivations.
However, deep graph learning does not leverage those insights.
Network science offers solutions to address open challenges in deep graph learning: principled data augmentation, rigorous evaluation practices, higher-order models, and temporal pattern recognition.
These solutions emerge by connecting network structure with function through principled methodologies, such as probabilistic generative models that provide principled null models for complex networks, statistical inference and network reconstruction methods for noisy relation data, and community detection techniques.
These approaches can enhance the theoretical foundation and empirical insight of deep graph learning models.

\textbf{Our position is that, without leveraging insights from network science, deep graph learning will stall.}
The past decade has seen rapid advances in deep graph learning architectures across various tasks and applications.
However, challenges to apply state-of-the-art graph neural networks to real-world problems also expose limitations that we need to address \cite{Georgousis2021,ju2024surveygraphneuralnetworks}.
We need data augmentation techniques that model noisy or incomplete data to improve generalization. We need theoretically grounded pooling methods that minimize information loss. We need systematic frameworks to characterize which structural properties of temporal graph datasets drive performance. And we need message-passing architectures that capture higher-order interactions beyond pairwise edges.
Insights from network science can help us to address these challenges, guiding deep graph learning toward principled architectures, improved interpretability, and more rigorous evaluation methods.

In this position paper, we formulate six \emph{Calls for Action} to integrate insights from network science into deep graph learning and bridging the scientific fields.
Ultimately, however, we believe that this is not a one-way road, and we aim to spur conversations across the two communities for mutual benefits.

\begin{figure*}
\centering
\includegraphics[width=.89\textwidth]{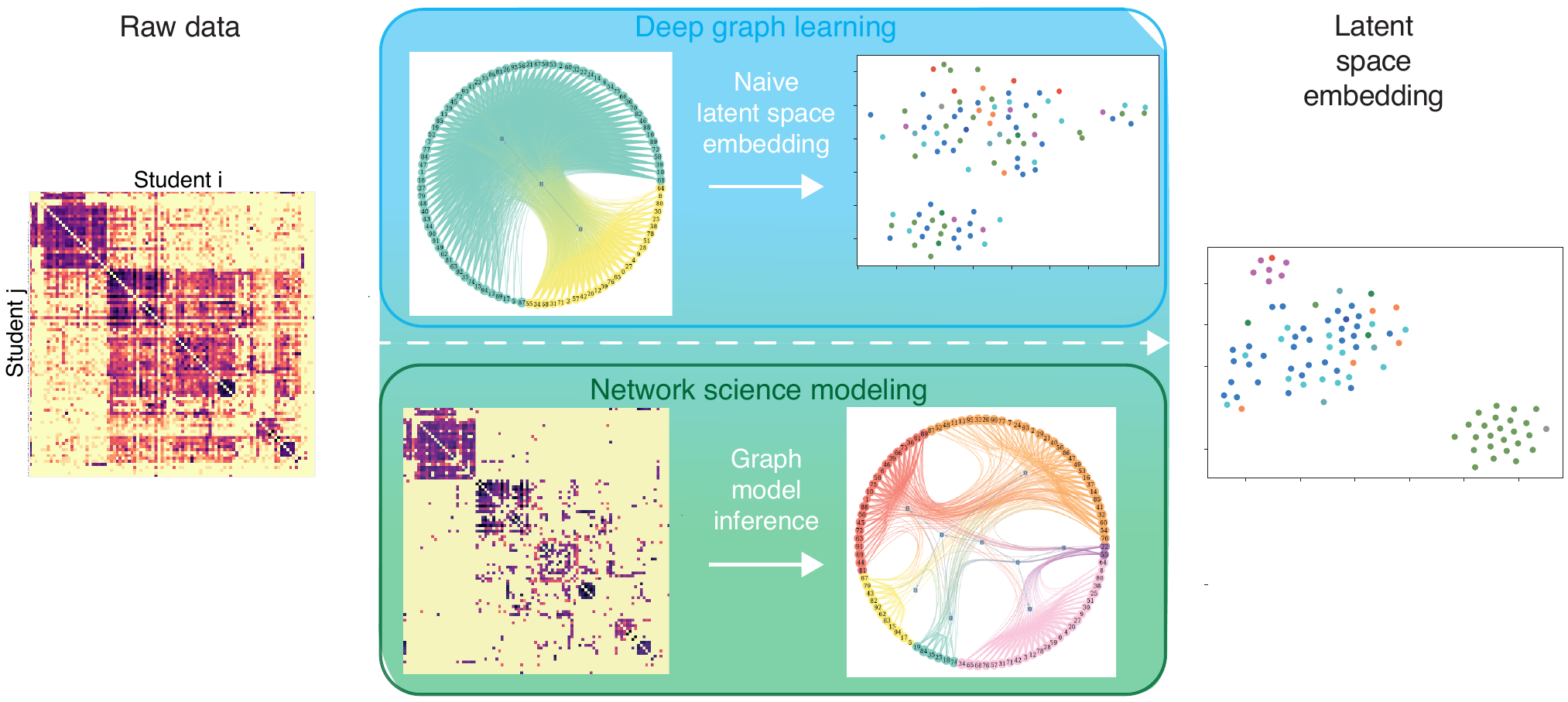}
\caption{\textbf{Illustrative example of how network science insights can advance deep graph learning}. Deep graph learning provides tools for end-to-end representation learning for prediction tasks (top panel), while network science provides advanced statistical modeling techniques for handling noisy graph data (bottom panel).
Applying deep graph learning directly to noisy co-occurrence data (left) results in a latent space representation that poorly reflects ground truth node labels (top panel). Network science modeling techniques---such as statistical ensembles of random graphs that preserve aggregate characteristics of the data---help build robust graph models that account for noise (bottom panel). Combining these techniques with deep learning methods leads to a latent space representation that better captures ground truth patterns (bottom right). Figure partly adapted from \cite{casiraghi2017}.\label{fig:example}}
\end{figure*}

\section{Probabilistic Models for Graphs}

Machine learning often uses \emph{data augmentation techniques} during training to improve model generalizability and mitigate overfitting.
This typically involves enriching available training data through perturbations, noise injection, or other means to augment it with artificial examples.
Inspired by these methods, the deep graph learning community has considered various \emph{graph augmentation} techniques that seek to manipulate edges or nodes of a graph in such a way that it improves the performance and/or generalizability of graph neural networks (GNNs) \cite{zhao2023graphdataaugmentationgraph}.
Recent works have considered, for example, targeted edge removal to increase homophilic patterns \cite{Zhao2020} or selectively adding nodes to slow down message passing \cite{azabou2023half}.
While these works on graph augmentation have made progress towards improving the generalizability of GNNs, researchers from the deep graph learning community have recently argued that we still lack a ``theory of data augmentation'' for graphs \cite{morris24a}, highlighting that, despite theoretical results on their expressivity, GNNs have not yet matured to leverage the full potential of the data in practical settings but currently rely on preprocessing for data augmentation.

It has also been pointed out recently that deep graph learning datasets suffer from serious limitations \cite{bechler-speicher2025position}:
The exact way in which a graph dataset was constructed determines its suitability for a specific prediction task.
However, modeling decisions are often not sufficiently grounded in theory, it is difficult to ensure that the patterns captured in a dataset are as intended, and in some cases, deep graph learning models may overfit to spurious patterns.
These issues underscore the need for explicit null models to safeguard against problematic data quality.

\textbf{Suggested Remedies.}
Network science can help us develop such a ``theory of data augmentation'':
Network scientists have long argued that it is often not desirable to use an observed network ``as is'' for analysis.
Many empirical complex network datasets are unreliable because they contain spurious or missing relationships, incomplete node sets, incorrect node labels, or noisy attributes.
For such data, network science has developed reconstruction and inference techniques \cite{peixoto2019network,newman2018network} that use statistical graph ensembles to infer reasonable graph models from noisy or incomplete data, as we illustrate in \Cref{fig:example}.
In deep learning, such methods can be used to generate a set of plausible graphs for training graph neural networks, thereby improving robustness and generalisability.

Random graph models are an important foundation of network science and define ``statistical ensemble''---all graphs that share given aggregate characteristics such as size, density, degree sequence or distribution, modular structure, or motif statistics.
These models can be used to randomize the topology of empirical networks while maintaining aggregate properties.
Such a randomization serves as a null model for statistical hypothesis testing, enabling us to understand which of a network's characteristics are due to the topology of the network, and which characteristics can be explained based on the mere degree distribution or edge density.
Important examples include the Erd\H{o}s-R{\'e}nyi random graph model where edges between pairs of nodes are randomly generated with equal probability \cite{erdos1960evolution}, the Molloy-Reed model that generates random graphs with a given degree sequence or distribution \cite{molloy1995critical}, exponential random graph models for random graphs with a given set of network statistics \cite{robins2007introduction}, or the stochastic block model for random graphs with given homophilic or heterophilic community patterns \cite{lee2019review}.
In network science, such statistical ensembles are the foundation to analytically study expected properties of graphs with given aggregate characteristics, for example, using generating functions as a framework \cite{newman2001random,newman2009random}.
Unfortunately, principled null models for graph-structured data are not yet widely used in the evaluation of deep graph learning architectures, where the focus often lies on the performance of a specific model in a given task rather than on which topological features of a graph can explain the predictive power of a given architecture.
Leveraging network science models can thus lead to more rigorous and meaningful evaluation practices that provide insights into the predictive capabilities of deep graph learning models.

\textbf{Call for Action 1:} Deep graph learning needs to adopt probabilistic models for graphs to address data uncertainty and express expectations about the data in a principled way.

\section{Principled Coarse-Graining Methods}

Pooling node representations into graph-level features, reducing graph complexity for computational efficiency, or identifying meaningful substructures require graph neural networks to compress information.
Graph pooling simplifies graphs by merging nodes into clusters, creating compact representations that improve standard applications, including classification and prediction.
Community detection identifies groups of nodes that are more densely connected internally than externally, providing a powerful lens to uncover meaningful structures in complex systems.
These coarse-grainings determine whether a method captures essential patterns or discards critical information.

Most graph neural networks approach coarse-graining in two steps.
First, they embed nodes into a continuous vector space.
Then they cluster these embeddings or apply pooling operations to create higher-level representations.
This pipeline places enormous pressure on the embedding step: it must capture all structure relevant to downstream tasks.

Neural embedding methods, such as node2vec, GraphSAGE, and autoencoder-based embeddings, demonstrate impressive capabilities in capturing community structure \cite{grover2016node2vec,hamilton2017inductive,kipf2016variational}.
These approaches often combine multiple components to construct and divide embeddings into communities.
For example, node2vec's random-walk approach can approximate spectral properties of the normalized Laplacian under controlled conditions, enabling it to approach the theoretical limits of community detection for certain graphs \cite{kojaku2024network}.
But this performance relies on a complex interplay between walk parameters, embedding dimensions, and subsequent clustering choices.
The opaque pipeline obscures which design choices drive performance and how to diagnose failures.

\textbf{Suggested remedies.}
Network science provides a well-established and interpretable framework for community detection.
Methods such as the stochastic block model (SBM) \cite{peixoto2019bayesian} and the map equation \cite{rosvall2008maps} have been developed and refined over years, each grounded in clear mathematical objectives.
The SBM identifies latent groups by modeling network topology.
The map equation highlights modular regularities in network flows, focusing on how information, resources, or behaviors propagate through the system.
Like node2vec, this flow-based approach proves valuable in applications where understanding dynamic processes matters, such as transportation networks, communication systems, and biological pathways.
Unlike opaque embedding pipelines, both Bayesian inference of the SBM and the map equation directly quantify how well different partitions compress the processes generating networks and flows on networks.
This transparency makes community assignments interpretable.

The rigour of network science extends beyond methodology to evaluation.
Generative models create benchmark networks with known ground truth.
Theoretical frameworks systematically compare community detection techniques, offering a foundation against which machine learning approaches can be measured~\cite{lancichinetti2008benchmark,bloecker2024flowdivergence}.
Incorporating these insights into machine learning evaluation workflows would strengthen the theoretical and practical reliability of community detection and pooling in GNNs.

Yet a challenge remains.
Network science builds on discrete mathematics: discrete objective functions, discrete data models, and combinatorial optimization algorithms.
Community detection traditionally considers hard partitions where each node belongs to exactly one community.
Algorithms optimize these partitions through discrete stochastic search, moving one node per iteration~\cite{Blondel_2008,edler2017infomap,Traag2019}.
Deep learning requires continuous, differentiable loss functions amenable to backpropagation and GPU-accelerated gradient descent optimisation.
Integrating discrete approaches from network science with gradient-based optimization requires making objective functions continuous and differentiable.

Recent deep graph clustering works have made this leap.
They adapted community detection approaches by treating community memberships as continuous rather than discrete, including the modularity criterion~\cite{tsitsulin2023dmon}, Poisson random process model~\cite{shchur2019overlapping}, minimum cuts~\cite{pmlr-v119-bianchi20a}, and the map equation~\cite{blocker2023map}.
Soft community assignments naturally produce overlapping communities that better capture real systems where nodes belong to several groups, such as in social networks.

These continuous adaptations represent an immense opportunity.
Principled unsupervised loss formulations can become part of composite losses to guide learning in supervised tasks.
Such methods can replace or complement purely data-driven pooling operations, providing more interpretable reductions~\cite{deng2024module,vonpichowski2024,castellana2025bnpoolbayesiannonparametricapproach}.
A challenge persists: continuous versions of such losses often require explicit regularization to avoid trivial solutions~\cite{pmlr-v119-bianchi20a,tsitsulin2023dmon,shchur2019overlapping}.

These promising developments lead us to formulate our second call for action:

\textbf{Call for Action 2:} We should incorporate principled and interpretable coarse-graining methods from network science to reliably capture meaningful structure in complex networks.

\section{Causality in Temporal Graph Learning}

Due to the growing availability of time-series data, the application of graph neural networks to temporal graphs, where edges and/or nodes change over time, has recently seen a surge of interest.
Many recent architectures are adaptations of existing deep learning methods that model (sequential) patterns in time-evolving batches of edges, which capture the time-varying topology of a temporal graph \cite{longa2023graphneuralnetworkstemporal}, including the event-based Temporal Graph Network model \cite{rossi2020} or snapshot-based approaches like the EvolveGCN architecture \cite{pareja2020evolvegcn}.
While the performance of these models has been evaluated for several tasks in empirical temporal graphs provided by, for example, the Temporal Graph Learning Benchmark (TGB) \cite{huang2023}, it is often unclear which patterns in temporal graphs they can actually learn.
In fact, recent systematic evaluations not only uncovered a complex picture of which (simple) temporal patterns state-of-the-art temporal GNNs actually learn \cite{hayes2025temporalgraphlearningmodels,su2025temporal}, they even yield the surprising insight that the time dimension does not seem to affect the performance of some temporal GNNs at all \cite{hayes2025temporalgraphlearningmodels}.

Hence, our understanding of what existing TGNN architectures actually learn is at best limited and relatively few works started to formally study the expressivity of temporal GNNs \cite{BeddarWiesing2024}.
Moreover, a major promise of including the time dimension as a ``first-class citizen'' in temporal graph learning is that it can facilitate causal deep learning in graphs, which is one of the grand challenges in artificial intelligence \cite{Schoelkopf_2022}.
For these reasons, temporal graph learning currently fails to deliver on this promise.

\paragraph{Suggested Remedies.} In network science, analyzing and modeling patterns in temporal networks have been topics of major interest for almost two decades \cite{holme2015modern}.
Network scientists have developed measures and models that capture different temporal, topological, and temporal-topological patterns found in real-world temporal graphs, along with generative models that selectively reproduce specific temporal patterns and destroy others.
Examples include measures and models that capture bursty activation patterns of nodes or edges \cite{moinet2015burstiness,takaguchi2013bursty,karsai2011small}, the temporal evolution of community patterns \cite{peixoto2017modelling}, as well as temporal motifs or correlations in the sequential order of interactions \cite{kovanen2011temporal,pfitzner2013betweenness,pan2011path}.
Selectively applying such generative models to empirical temporal graph data allows us to ``disentangle'' temporal, topological, and temporal-topological patterns.
However, few works have applied them to systematically understand which types of patterns existing temporal GNNs can actually learn \cite{hayes2025temporalgraphlearningmodels}.
Much work remains to be done in adapting existing network science models to understand which of those patterns are actually captured by a specific deep temporal graph neural network architecture.

Considering how the timing and ordering of temporal events affect models of dynamical processes, and, thus, how nodes can causally influence each other over time, network science has developed modeling approaches that capture sequential patterns in the ordering of time-stamped edges that influence time-respecting paths in temporal networks.
In a nutshell, two time-stamped edges $(a,b;t)$ and $(b,c;t')$ in a continuous-time temporal graph can form a time-respecting path connecting node $a$ via $b$ to $c$ if and only if $t < t'$, that is, if edge $(a,b)$ occurs before edge $(b,c)$.
Hence, due to the arrow of time, whether $a$ can possibly causally influence node $c$ depends on whether a time-respecting path exists.
Importantly, the existence of time-respecting paths is only a necessary but not a sufficient condition for the presence of causal influence.
Nevertheless, neglecting this necessary condition hinders the development of a new generation of temporal graph learning methods that account for causal effects.

Network science has developed approaches to represent, model, and analyze patterns in the time-respecting path structure of temporal graphs, which is sometimes called causal topology or causality structure.
Examples include directed acyclic graph representations \cite{Saramäki2019}, higher-order memory networks \cite{rosvall2014memory}, as well as high-dimensional De Bruijn graph models \cite{scholtes2014causality}.
Recent works in deep learning have successfully adopted these ideas to derive temporal GNNs that account for the causal topology of temporal graphs, demonstrating improved performance across several temporal graph learning tasks such as link prediction, node classification or regression~\cite{WangCLL021,qarkaxhija2022,heegusing}.
Providing a foundation for studying how time-respecting paths influence temporal graph learning methods, \cite{Stamm2025} recently proposed an interesting new sampling technique that preserves the causality structure of a given temporal graph, which can help us investigate which of the existing temporal GNNs can actually learn such patterns.
The citations in these works show that that they are are directly inspired by network science works modeling time-respecting paths in temporal graphs, thus demonstrating how network science insights and models can advance temporal graph learning.
The promising results from the approaches that bridge those views motivate our third call for action:

\textbf{Call for Action 3:} We must better incorporate the rich body of network science findings on relevant patterns in temporal networks to leverage the full potential of time series data to advance graph learning.

\section{Understanding Neural Message Passing}

Message passing is a core mechanism in GNNs \cite{Hamilton2020}.
It allows nodes to gather information from their neighbors to update their representation, supporting tasks such as link prediction and node classification.
While the specifics regarding message, update, aggregation, and activation functions differ between GNN architectures \cite{ju2024surveygraphneuralnetworks}, most of them follow similar principles -- and suffer from similar limitations such as, for example, poor performance for networks with heterophilic patterns \cite{zhu2020beyond}, the insensitivity of node representations to information from distant nodes (``oversquashing'') \cite{di2023over}, or the exponential convergence of node representations with increasing number of message passing known as ``oversmoothing'' \cite{rusch2023survey}, or the fact the performance of neural message passing architectures is sometimes surpassed by simple heuristics or neural network architectures that ignore the graph topology altogether \cite{NEURIPS2023_0be50b45,hu2021graph}.

\paragraph{Suggested Remedies.}
Understanding the interplay between the structure of complex networks and the evolution of dynamical processes like, for example, epidemic spreading, synchronization, Laplacian diffusion, or percolation has been a key focus in network science for more than two decades~\cite{boccaletti2006complex,barrat2008dynamical}.
This research has resulted in a variety of structural measures and analytical frameworks that can help us understand message passing and, in particular, which types of complex networks neural message passing succeeds in and which it fails in.
As an example, the often remarkable performance of GNNs may be seen with different eyes, if we consider that, even without any trainable parameters, the ``pure'' message passing rule of popular GNN architectures like GCN, GraphSAGE or GIN effectively calculates node similarity scores.
For example, a single iteration of pure message passing aggregates information from neighboring nodes, enabling a GNN to approximate the number of common neighbors shared by two nodes \cite{dong2024pure}.
Similarly, multiple layers of (untrained) neural message passing have been shown to capture more complex, path-based similarity patterns, as measured by rooted PageRank, the Katz index, the Adamic-Adar index, or SimRank \cite{qarkaxhija2025}.

These works highlight a promising connection between GNNs and network science, where network science provides new ways to analyze neural message passing architectures.
Constituting another concrete example, in network science probability generating functions have been developed as a powerful analytical framework to study various phenomena in complex networks, for example, the friendship paradox in social networks, the dynamics of epidemic spreading, or the critical percolation threshold in generalized random graphs with arbitrary non-Poissonian distributions \cite{newman2001random}.
In a nutshell, this framework elegantly derives the distributions of quantities, such as the number of reachable nodes, sampled by traversing random graphs up to a given depth.
\cite{lampert2024} adopt this framework to gain insights into deep graph learning, specifically studying the influence of self-loops in message passing neural networks like GCN
\cite{hamilton2017inductive}.
Many GNNs include self-loops, since they allow incorporating the features of a node itself in the aggregation step, omitting the need for an explicit update function \cite{Hamilton2020}.
Intuitively, one would expect the inclusion of self-loops to strengthen the influence of a node's own features on its learned representations.

Assuming only knowledge of a graph's degree sequence, applying the framework of \cite{newman2001random} allows us to analytically calculate the expected number of walks between specific pairs of nodes.
This allows us to prove the ``self-loop paradox'', the counterintuitive finding that, in a graph neural network with two message-passing layers, the inclusion of self-loops can actually \emph{decrease} the impact of a node's own features on the representation of that same node \cite{lampert2024}.

The suggested remedies above lead us to formulate our fourth Call for Action:

\textbf{Call for Action 4:} We should embrace network science methods to improve the interpretability of deep (graph) learning models.

\section{Universal Organization Principles of Complex Networks}

Graph neural networks are complex and seemingly unintelligible. They are often viewed as "black boxes" and treated as static parameter matrices rather than graphs shaped by a set of universal organizing principles. In these networks, as they are in many other networks, a small number of nodes, edges, or attention heads dominate much of the dispersion of message passing across these networks. This partially explains why many GNN models are robust to random noise, while at the same time, sensitive to specific kinds of network ablations. But we still lack the ability to predict, for example, when these networks are likely to be robust and when they will fail.

Foundation models have similar, transcendent properties that cut across any one specific graph, while at the same time exhibiting flexibility to varied contexts and domains. In natural language processing, we see cross-lingual transfer of LLMs (i.e., between high-resource and low-resource languages), enabled by the ability of LLMs to model \emph{universal language patterns} from individual sets of languages. These universal patterns are partly due to joint genealogies of languages and---for languages that evolved independently---similar evolutionary principles. But our understanding of the mechanisms that drive these "universal" patterns is lacking.

So, how do we better predict failures or explain these seemingly universal principles?

\textbf{Suggested Remedies.} Network science has spent decades investigating the universal organizing principles of complex networks. Most real-world networks in biology, sociology, and economics share similar architectural properties and constraints. They are sparse but highly structured, hierarchical with dense cores and sparse periphery, heavy-tailed, homophilous, and redundant with overlapping pathways. The dynamics on these networks (e.g., flow of information or disease spread) shape their structure and structure shapes the dynamics. Initial conditions (e.g., path dependence) have downstream effects on these networks, and trade-offs (e.g., efficiency versus robustness) help explain ensemble improvements, brittleness, and the convergence structures that we often see in real-world networks, and also in graph neural networks.

These universal principles of networks can serve as useful guideposts for better understanding the properties of GNNs. As noted above, the disproportionate influence of a small percentage of nodes is something observed and studied extensively in network science and something we see now with GNNs. Network structures also often reflect optimization processes under constraints, rather than simply random processes. This could mean, at least for GNNs and similar graphs, that they are not simply "black boxes" but \textit{legible} and interpretable if we correctly identify those constraints.

Network science can partially explain why GNNs are trainable. The tight-knit communities and strong homophily found in many graphs studied in network sicnece also have short paths, which enable fast diffusion, even with the deepest, most complex neural networks.

Network science has been grappling with the fascinating observation that complex networks, whether in an individual human cell or at the level of human society, share similar, universal properties---from small-world patterns to preferential attachment processes.  In other words, network science provides the kind of theoretical constraints that could helpful in explaining the limits, failure modes, and the surprisingly useful properties of GNNs.

In the following three sections, we will elaborate on three additional areas where knowledge of universal principles of complex networks could help advance deep graph learnings.

\subsection{From Example-based to Unsupervised Learning}
Deep graph learning methods focus on learning data patterns in a supervised, semi-supervised, or self-supervised fashion.
Supervised and semi-supervised learning rely on labeled data to measure performance and train models' weights via backpropagation to minimize training loss.
Consequently, they require their training data to contain examples covering all labels that will be considered during prediction.
If this is not the case, the trained model cannot identify instances belonging to the unseen label.
A way around this is to apply self-supervised techniques such as graph auto-encoders that encode and then decode the data, compressing it to a lower-dimensional representation, followed by decompressing it to reconstruct the original data \cite{kipf2016variational,9770382}.
The training criterion in this case is to minimize the discrepancy between original and reconstructed data, which can be measured via the $L_2$ norm or other measures that quantify the ``distance'' between matrices.
However, this approach allows the user very limited control over the exact patterns that are learnt during the embedding process.

In contrast, network science methods typically do not assume labeled training data and employ models designed to capture specific patterns seen across networks of varied size, structure, and type.
This makes them more widely applicable in a world where most data is unlabeled and research questions require focusing on a specific aspect of the data in a controllable way.
Integrating such model-based approaches from network science with deep graph learning methods would make it possible to align GNNs with transparent and interpretable models and train them in an unsupervised fashion.

\subsection{Balancing Interpretability and Flexibility}
Deep graph learning and network science take different approaches to modeling graph data.
Network science provides interpretable frameworks based on explicit assumptions about structure and relationships, but these assumptions can limit flexibility in complex graphs.
In contrast, deep learning methods prioritize flexibility, accommodating diverse inputs and intricate dependencies, often at the cost of interpretability.
This trade-off between interpretability and flexibility is an opportunity for deep graph learning to incorporate insights from network science while maintaining flexibility.
For example, GNNs struggle in heterophilic settings, where connected nodes differ in features or labels \cite{zheng2024what}.
In contrast, statistical models like the SBM can capture both homophilic and heterophilic structures.
Integrating such principled models into GNNs could improve generalization and robustness across varying graph topologies.

Meta-learning emerges as a promising approach to balance interpretability and flexibility.
It enables models to dynamically adjust their assumptions or learning strategies based on characteristics of the input data.
Hybrid frameworks can combine the interpretable structure of probabilistic network models with the scalability and flexibility of deep graph learning.
For example, integrating automatic differentiation and Bayesian inference enables scalable, interpretable models capable of handling diverse network configurations \cite{contisciani2025flexible}.
Similarly, GNNs inspired by structured models like the SBM could better capture heterophilic patterns, improving their applicability across different graph types~\cite{pmlr-v235-wang24u}.

\subsection{Towards Foundation Models for Graphs}
Finally, a major open challenge in deep graph learning is that current tasks are still largely domain-centered and models need to be trained for a specific task in a given graph or, for graph-level learning tasks, a specific set of graphs from a given domain.
Different from recent advances in computer vision and natural language processing, deep graph learning still largely lacks \emph{foundation models} that could generalize to new, previously unseen graphs \cite{morris24a}.
While deep graph learning has now taken the first steps in this direction \cite{zhao2024graphanyfoundationmodelnode,Haitao2024}, network science has a long tradition of identifying and modeling \emph{universal organizing principles} that govern the structure and evolution of networks across domains like social networks, information systems, biology, and large-scale infrastructures.
To this end, many insights have been generated regarding how simple (local) mechanisms shape collective patterns and characteristics of large-scale networks and/or dynamical processes across domains.
Concrete examples include phase transitions in the connectivity of large graphs that can be explained based on the ratio between moments of their degree distributions \cite{newman2001random}, simple growth rules like preferential attachment that lead to similar scale-invariant patterns in very different networks \cite{barabasi1999emergence}, or common characteristics in financial and social networks that result in similar propagation and consensus dynamics across different systems \cite{lorenz2009systemic}.
Despite these strong results in network science, we lack a theoretical understanding of whether and how these common organization principles in networks from across different domains affect the generalization capabilities of commonly used deep graph learning architectures, and how we could build architectures able to capture such principles.

\textbf{Call for Action 5:} We should learn from network science's insights into universal organization principles of complex networks. In doing so, we can move beyond "black boxes" to graphs optimized and constrained by a set of universal, evolutionary principles.

\section{Bridging Scientific Cultures}

Network science and deep graph learning address similar problems.
The complex graphs found in both fields are fascinatingly flexible and robust across varied contexts, yet also brittle and difficult to interpret. Similar evolutionary principles and architectural trade-offs help explain some of these universal properties, yet the fields operate largely independently.
Part of this separation stems from differences in scientific culture.
Network scientists publish mainly in interdisciplinary journals or in outlets in statistical physics, whereas the deep learning community focuses on a small set of major machine learning conferences.
This divide makes communicating results difficult and affects the strategies junior researchers adopt.
Since securing tenure depends on publication venue prestige, junior researchers face incentives to remain within established communities rather than bridge them.

The two communities also use different terminology, sometimes for the same concepts.
Many concepts in network science, such as statistical ensembles, phase transitions, and generating functions, originated in statistical physics.
These terms carry precise meanings for network scientists with physics backgrounds, but remain unfamiliar to most machine learning researchers with backgrounds from computer science.

These barriers reduce visibility in both directions.
The large body of network science results in physics journals remains largely unknown to the deep learning community, but also vice versa.
Publishing cultures limit the dissemination of deep learning work that uses network science models.
The network science community sometimes views the rapid growth of deep graph learning as a competitive threat rather than an opportunity for collaboration.
The motivation for writing this perspective is to share actionable insights across the two fields and to begin bridging them.

\textbf{Suggested Remedies.}
Strengthening this connection requires bringing researchers from different backgrounds and publication cultures together.
Dedicated journal-first tracks at deep learning conferences could provide network science researchers with pathways to machine learning venues.
Workshops and special journal issues that explicitly bridge the two fields create forums for exchange.

\textbf{Call for Action 6:} We must bridge the scientific cultures between deep graph learning and network science through joint publication venues and forums for collaboration.


\section{Alternative Views}

Given recent advances in research fields adjacent to deep graph learning, especially the application of transformers \cite{transformer} in large language models (LLMs), the question arises whether deep graph learning will really stall without insights from network science.
While the debate over whether LLMs merely perform sophisticated text completion or truly have reasoning capabilities continues \cite{chi2024unveiling,kalai2025languagemodelshallucinate,yao2023tree}, there is evidence that they can learn grammatical concepts \cite{brinkmann2025largelanguagemodelsshare}.
From a graph-learning perspective, ``grammar'' could be seen as corresponding to universal organization principles of complex networks.
Hence, graph transformers \cite{dwivedi2020generalization} could potentially discover models and universal organization principles of complex networks as needed to solve the prediction task at hand.
Yet an important question arises: would we recognize that LLMs discover grammar if grammarians had not already identified grammatical structures? Similarly, can we claim graph transformers discover universal mechanisms without network scientists having studied them? And graph data is different from text.
Whereas text is abundant and grammatical concepts transcend domains, graph datasets are focused on a single specific problem, data-collection procedures differ, and resulting datasets vary in quality \cite{ogb,huang2023,bechler-speicher2025position}.
Can deep graph learning infer universal principles from such single-purpose datasets of varying quality?
Possibly, but network science has specifically focused on identifying universal principles, and we believe that deep graph learning should make use of all the lessons network science has learned.

Also related to transformers, one could argue that it will soon no longer be necessary to explain message-passing architectures, because they will be replaced by graph transformers, and thus, lose relevance.
Graph transformers could help avoid issues like oversquashing and oversmoothing, but we believe network science will remain relevant for studying the adaptive structure aggregation learned by the attention mechanism.
A testimony to this are recently proposed methods to reveal the learned computational graphs of transformer-based language models \cite{ameisen2025circuit}.
The application of network science methods to these computational graphs could enhance the mechanistic interpretability of black-box models.
It will also be interesting to see whether known organization principles and models of social, biological, and physical networks also apply to these learned computational graphs.

\section{Conclusion}

In this position paper, we argue that deep graph learning will stall without integrating insight from network science.
We highlighted several core challenges in deep graph learning and formulated Calls for Action to offer a pathway for integrating methods and insights from network science into deep graph learning.
By bringing the theoretical frameworks of network science to deep learning, researchers in the deep graph learning community can address challenges such as robust and principled data augmentation, effective pooling techniques, and higher-order modeling of complex interactions.

To realize this potential, we call for integrating network science methods---such as random graph theory, generating function analysis of random graph ensembles, and models of dynamical processes like diffusion, synchronization, or consensus dynamics---into computer science and deep learning academic curricula.
However, we do not see this as a one-way road and believe that network science and data science curricula would benefit from incorporating deep learning methods.
Ultimately, we aim to spur conversations across the two communities for mutual benefits:
At the scholarly level, fostering collaboration between network science and deep graph learning is essential.
Strengthening this connection requires bringing researchers from different backgrounds and publication cultures together, whether through dedicated journal-first tracks in deep learning conferences or by organizing workshops and special journal issues that bridge the two fields.
We see such a convergence as more than just a set of technical advances.
By embracing the interpretability of network science and the flexibility of deep learning, researchers from both fields will benefit---just as Hopfield did by integrating ideas across statistical physics, neuroscience, and complex systems.

\section*{Acknowledgements}
MR was supported by the Swedish Research Council under grant 2023-03705.
This work was partially supported by the Wallenberg AI, Autonomous Systems and Software Program (WASP) funded by the Knut and Alice Wallenberg Foundation.

\end{document}